\documentclass[letterpaper, 10 pt, conference，]{ieeeconf}  
\usepackage{amsmath}
\usepackage{graphicx}
\usepackage{tabularray}
\usepackage{txfonts}
\usepackage{amssymb}
\usepackage{marvosym}
\usepackage[noadjust]{cite}  
\makeatletter
\let\NAT@parse\undefined
\makeatother
\usepackage[colorlinks,linkcolor=red,citecolor=green]{hyperref}

\usepackage{epstopdf}
\usepackage{titlesec}
\titlespacing\section{2pt}{2pt}{2pt}
\titlespacing\subsection{2pt}{2pt}{2pt}
\titlespacing\subsubsection{2pt}{2pt}{2pt}

\AppendGraphicsExtensions{.tif}
\IEEEoverridecommandlockouts                              

\title{\LARGE \bf
TDCNet: Transparent Objects Depth Completion with CNN-Transformer Dual-Branch Parallel Network
}

\author{Xianghui Fan$^{1,2}$, Chao Ye$^{3,4}$, Anping Deng$^{1}$, Xiaotian Wu$^{5}$, Mengyang Pan$^{1}$ and Hang Yang\textsuperscript{\Letter}$^{1}$
\thanks{This work was supported by National Natural Science Foundation of China (NSFC) [grant number 62175086] }
\thanks{$^{1}$Changchun Institute of Optics, Fine Mechanics and Physics, Chinese Academy of Sciences, Changchun $130033$, China. \textsuperscript{\Letter}Corresponding author:
        {\tt\small yanghang@ciomp.ac.cn}}%
\thanks{$^{2}$University of Chinese Academy of Sciences, Beijing 101408, China}%
\thanks{$^{3}$Research Institute of Intelligent Control and Systems, Harbin Institute of Technology, Harbin $150001$, China}%
\thanks{$^{4}$National Key Laboratory of Complex System Control and Intelligent Agent Cooperation, Harbin Institute of Technology, Harbin $150001$, China}%
\thanks{$^{5}$College of Physics, Northeast Normal University, Changchun $130024$, China}%
}

\begin{document}

\maketitle
\thispagestyle{empty}
\pagestyle{empty}

\begin{abstract}

The sensing and manipulation of transparent objects present a critical challenge in industrial and laboratory robotics. Conventional sensors face challenges in obtaining the full depth of transparent objects due to the refraction and reflection of light on their surfaces and their lack of visible texture. Previous research has attempted to obtain complete depth maps of transparent objects from RGB and damaged depth maps (collected by depth sensor) using deep learning models. However, existing methods fail to fully utilize the original depth map, resulting in limited accuracy for deep completion. To solve this problem, we propose TDCNet, a novel dual-branch CNN-Transformer parallel network for transparent object depth completion. The proposed framework consists of two different branches: one extracts features from partial depth maps, while the other processes RGB-D images. Experimental results demonstrate that our model achieves state-of-the-art performance across multiple public datasets. Our code and the pre-trained model are publicly available at https://github.com/XianghuiFan/TDCNet.

\end{abstract}

\section{INTRODUCTION}
Depth perception is a fundamental component of robot perception. Normal active depth sensors, like LiDAR and other TOF cameras, rely on the Lambertian assumption, which assumes uniform light reflection from object surfaces in all directions. However, transparent materials do not conform to this assumption, making conventional depth sensors unable to accurately measure the surface depth of transparent objects \cite{review}. This also affects image-based techniques like stereo matching or monocular depth estimation, which tend to estimate the depth of the object or background behind transparent surface rather than the transparent surface itself. The false perception of transparent objects further creates difficulties for downstream applications, including robotic manipulation.
   \begin{figure}[tp]{
      \setlength{\fboxsep}{0pt}%
      \setlength{\fboxrule}{0pt}%
      \centering
      \framebox{\includegraphics[width=0.95\linewidth]{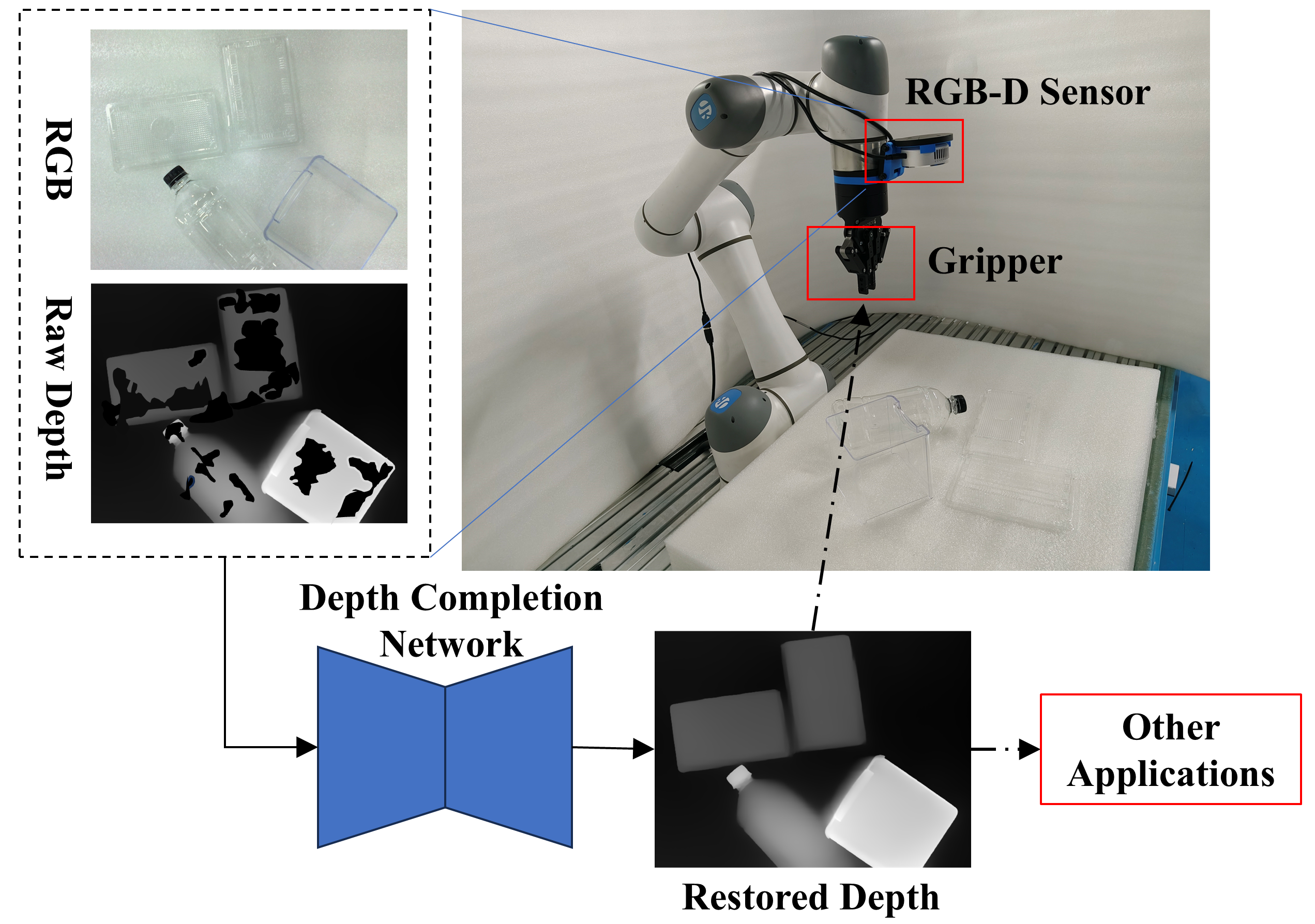}}

}
      
      \caption{
When encountering transparent objects, the depth map directly obtained from an RBG-D sensor is often incomplete, which poses challenges for robotic operations. The depth completion network can address this issue by reconstructing the incomplete depth map, enabling downstream applications such as robotic grasping or other manipulation tasks.
      }
      \vspace{-0.8cm}
      \label{figure1}
   \end{figure}
   
To address the problem of depth perception of transparent objects, some methods attempt to repair the original depth map acquired directly from the depth sensor. For example, ClearGrasp\cite{cg} uses a global optimization-based approach to repair depth maps, requiring only the original RBG-D image as input to obtain better depth prediction than monocular depth estimation. However, ClearGrasp needs to train three networks to predict surface normals, edges, and transparent object masks, which brings additional training costs. In order to reduce the cost, many studies\cite{transcg,implicit,fdct,tode,tcrnet,joint,segment,tim} have focused on using only one neural network to restore the depth of transparent objects from raw RBG-D images, which is also known as end-to-end methods. 

\begin{figure}[tp]{
   \vspace*{0.25cm}
      \setlength{\fboxsep}{0pt}%
      \setlength{\fboxrule}{0pt}%
      \centering
      \framebox{\includegraphics[width=\linewidth]{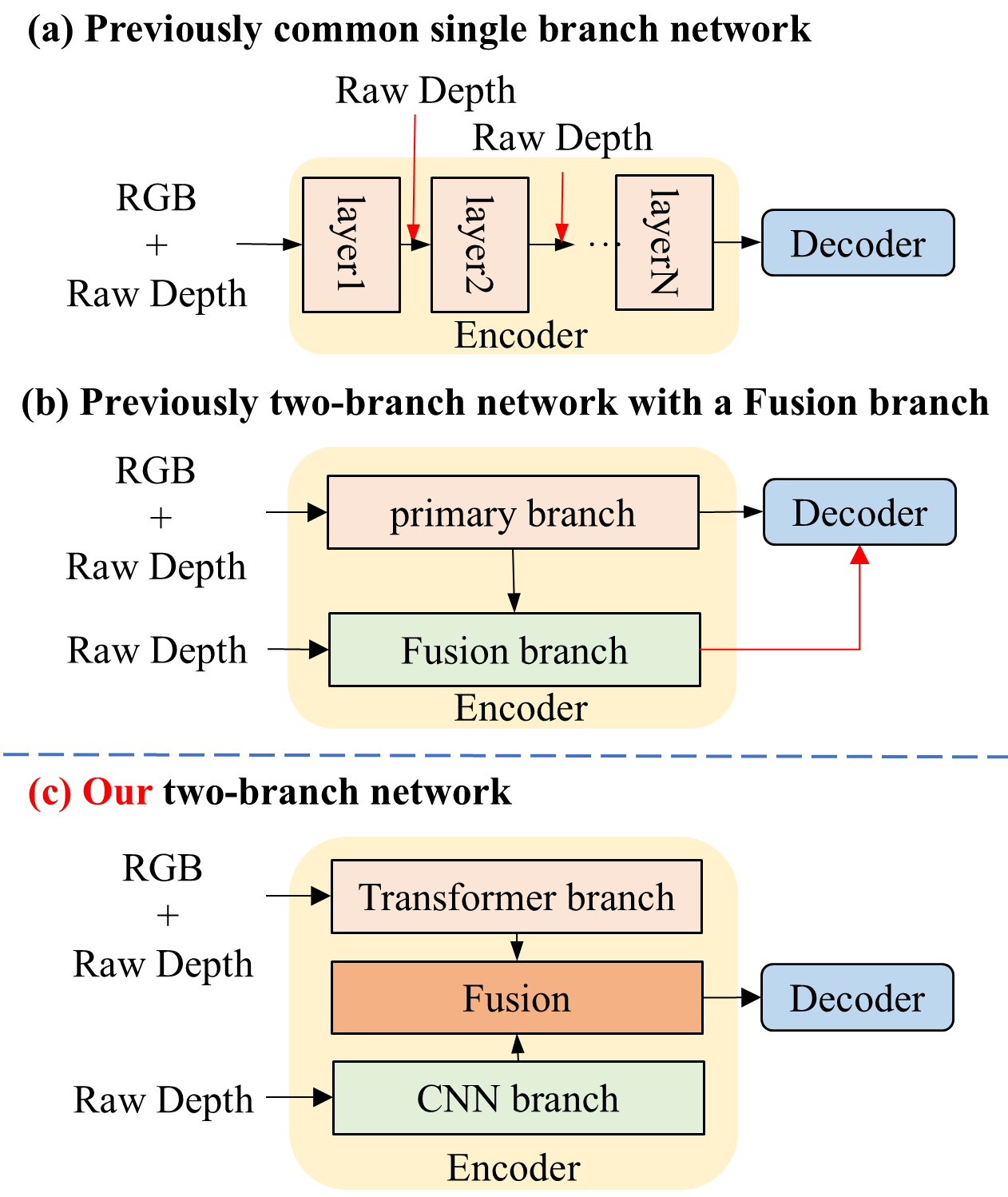}}

}
      
      \caption{
Comparison of the popular architectures for transparent object depth completion. \textbf{(a)} Previous common single-branch structure where depth maps are usually added in the middle layer. \textbf{(b)} Previous dual-branch structure with a fusion branch, where the fusion branch is used to fuse features from the original depth maps. \textbf{(c)} Our parallel dual-branch structure, where two branches with different backbones extract the features from the original depth maps and the RBG-D image, respectively, and then fuse them.
      }
      \label{figure2}
   \end{figure}

Unlike multi-branch designs for multi-source image fusion tasks\cite{imagefusion}, end-to-end methods in depth completion task usually design a network with a single-branch structure and use the RBG-D image as the input, e.g., \cite{transcg,fdct,tode,tcrnet,segment,tim}. This input results in low-level features from depth being blended with RGB at shallow layers of the network, whereas these features are important for the depth completion since they contain a large number of reliable information such as the value of depth, edges, and angles of transparent objects\cite{depthimportance1,depthimportance12,segment}. As shown in Fig. \ref{figure2} (a), some single-branch methods\cite{segment,transcg,tim} use layer-by-layer addition of the original depth map to compensate for middle layers, but this approach is sensitive to depth map noise and lacks feature extraction process for the depth map. 

Some methods\cite{swindrt}\cite{fdct} use an additional branch to extract features from the original depth map; e.g., FDCT\cite{fdct} uses a dual-branch network with a fused branch; Fig. \ref{figure2} (b) shows the structure sketch. The fusion branch combines features from both the main branch and the depth map, thereby reducing sensitivity to depth noise and enhancing the network's feature representation through feature-level fusion. However, the main branch dominates the whole network, and this branching imbalance undoubtedly weakens the influence of the fusion branch, further leading to a weakening of the influence of the depth features present only in the fusion branch.
   
In order to better utilize the features of the original depth map, we propose a CNN-Transformer parallel dual-branch network(TDCNet), as shown in Fig. \ref{figure2} (c), where our dual-branch uses the RBG-D image and the depth image as inputs, respectively. Unlike previous approaches, our dual-branch extracts features independently to prevent the blending of the original depth map's features. In addition, our fusion structure performs a fair fusion of the features from the two branches, thus forcing the model to focus on the features from the original depth map. We note that depth map features are always locally correlated, e.g., depth values and edges are always locally continuous, and CNN are known to excel at extracting local features. In addition, \cite{tode} has shown us that the single-branch Transformer architecture is still highly performant in the dense prediction. The respective advantages of CNN and Transformer have given rise to our parallel CNN-Transformer backbone design. Additionally, due to the coexistence of multiple feature modelling methods, using different backbones simultaneously has the advantage of richer feature representations.

In addition, we design a fusion structure for fusing dual-branch features and multiscale features. In this structure, we use the Multiscale Feature Fusion Module (MFFM), which accepts two features from neighbouring scales as inputs, then computes weights from multiple dimensions based on the inputs. Our ablation experiments demonstrate that our module is more effective than methods that do not perform multi-scale feature fusion or non-learning fusion.

Finally, we note that previous methods usually use a combination of primary and auxiliary loss functions for training, which inevitably result in the gradient conflict problem. We draw on the experience of multi-task learning \cite{MTLDeacy,MTLRate} to suppress the weights of the auxiliary losses after training as convergence is approached, so as to allow the model to be more focused on the depth recovery in the later stages of the training. Our experiments demonstrate that our strategy is more effective compared to the previous method that used a fixed weight combination.

Our main contributions can be summarised as follows:
\begin{itemize}
\item We propose a CNN-Transformer parallel dual-branch network for the transparent object depth completion task. The network uses different backbone designs for dual-branch. We designed a large number of experiments to demonstrate the effectiveness of this design.
\item We designed a fusion structure for fusing dual-branch features and multi-scale features. The core of this is the MFFM module for multiscale fusion. We also design a training strategy for suppressing gradient competition due to the combination of multiple loss functions.
\item We evaluate our model on several commonly used public datasets, and the experimental results show that our model achieves state-of-the-art performance and has excellent generalization ability.
\end{itemize}

\section{RELATED WORKS}
   \begin{figure*}[htbp]
   \vspace*{0.25cm}
      \setlength{\fboxsep}{0pt}%
      \setlength{\fboxrule}{0pt}%
      \centering
      \framebox{\includegraphics[width=\linewidth]{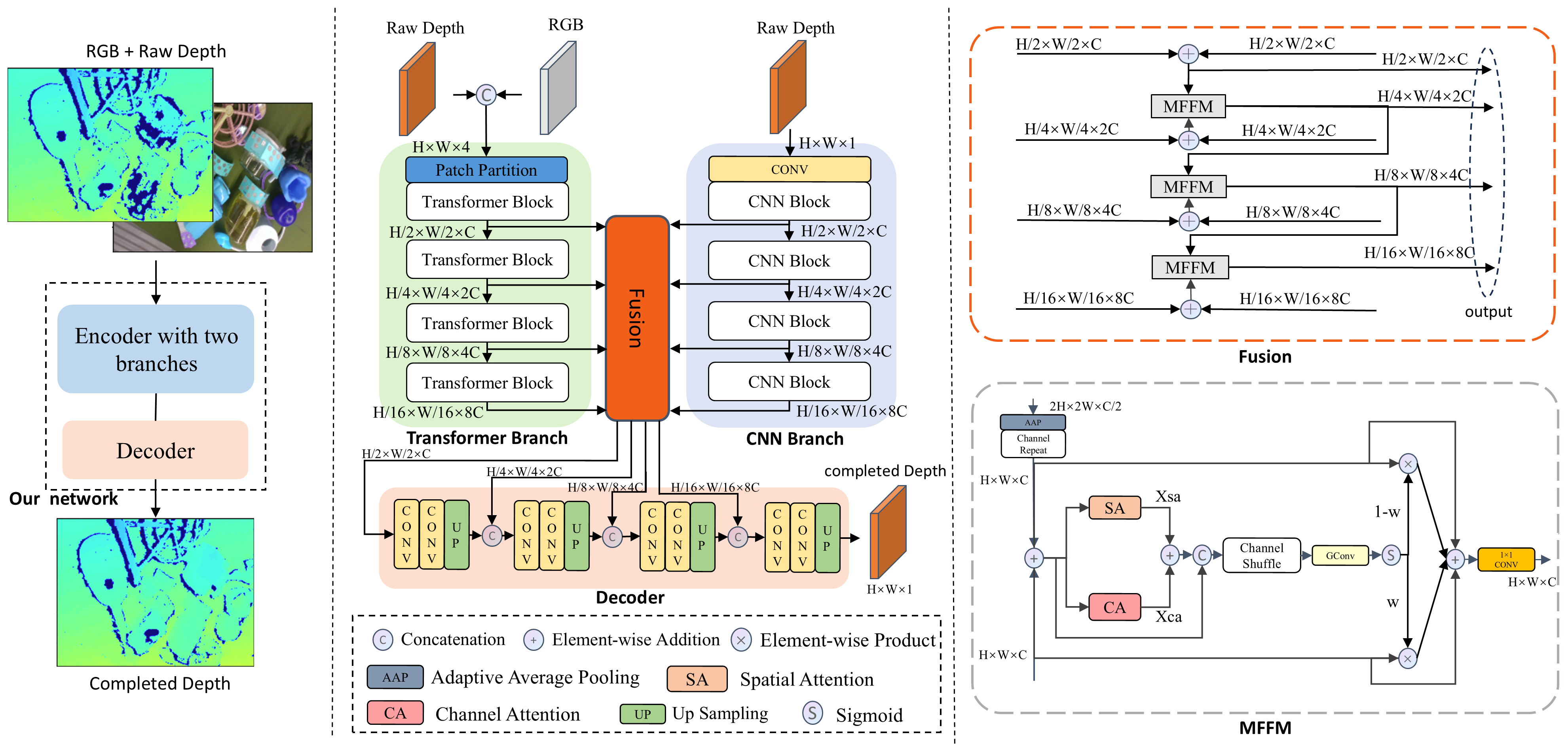}

}
      
      \caption{
      The architecture of TDCNet. Our network consists of two parts: an encoder and a decoder. The encoder consists of two parallel branches and a fusion structure. The two branches use CNN-based and Transformer-based backbones to extract the original depth map and RBG-D (4 channels) features, respectively, and our fusion structure based on the MFFM collects and fuses the features from the two branches from multiple scales. The decoder comprises full convolution modules and upsampling modules, which ultimately process the encoder's features to produce the final depth map.
      }
      \vspace{-15pt}
      \label{network}
   \end{figure*}
\subsection{Depth Completion for Single View Transparent Objects}
How to quickly and accurately complement the depth of a transparent object from a single viewpoint is a hot research topic since single views can be easily acquired. Some early methods rely on external assumptions, such as backgrounds\cite{out_tech1,out_tech2}, and sensing devices\cite{out_tech3,out_tech4}, making them inapplicable to randomized scenes. Some methods\cite{cg,a4t} use the global optimization algorithm\cite{global_method} to overcome the limitation of external assumptions, and only need the RBG-D image acquired by the sensor as input to complement the depth of the transparent object. However, both methods in \cite{cg,a4t} rely on the prediction of surface normals and edges, and to overcome this dependency, recent research has been devoted to using only an end-to-end model for the depth completion task. For example, in Zhu \textit{et al}\cite{implicit}, use a local implicit neural representation and an iterative depth refinement model to complement the depth of a transparent object. Xu et al. \cite{joint} propose a method called transparentNet, which simultaneously complements both the point cloud and the depth map of a transparent object. In \cite{depthgrasp}, a generative adversarial network is used to generate a depth map for reconstructing transparent objects. In \cite{transcg}, a multiscale depth network is proposed that connects RGB and depth maps and uses them as inputs. Both ToDE-Trans\cite{tode} and swinDRNet\cite{swindrt} use swin Transformer. ToDE-Trans uses an encoder-decoder structure, while swinDRNet designs a two-stream fusion network and uses the predicted confidence maps to fuse the original and predicted depths. In Zhai \textit{et al}\cite{tcrnet},  designed a decoder consisting of multiple modules called refinements, which are connected to each other in a cascade fashion. For feature fusion between different layers and low-level depth feature exploitation, Li \textit{et al}\cite{fdct} added a fusion branch with a hopping structure to the Unet structure. Our model follows the end-to-end design and encoder-decoder structure, and different from previous two-stream fusion networks, we design separate CNN branches for the original depth map and fuse them with RBG-D features extracted from another branch, and use a learnable module to fuse multiscale features.

\subsection{CNN-Transformer Hybrid Architecture for Vision Tasks}
ViT\cite{vit} models have excellent performance on specific computer vision tasks, and this success can be attributed to the strong global modeling capability of the mechanism of Multi-Head self-attention(MSA)\cite{msa}. However, since ViT lacks the ability to inductively bias local relationships, their ability to generalize to certain applications is limited. Some studies have proposed to solve this problem by integrating convolutional and self-attention mechanisms, which fall into three categories: sequential, parallel, and chunked integration\cite{CNN-Transformersurvey}. In contrast to \cite{CNN-Transformer-example1} which capture local patterns with CNNs and then learns long-term dependencies with MSA, Li \textit{et al}\cite{CNN-Transformer-example2} use CNN in the late stage of the model and MSA in the early stage. In \cite{CNN-Transformer-example3}, researchers alternately employed convolutional mechanisms and MSA. Parallel use of CNN and Transformer is common on specific tasks, e.g., \cite{CNN-Transformer-example4}, where the common practice is to use CNN and Transformer branches separately to extract features, and then perform feature fusion at a late stage. Some works, such as \cite{CNN-Transformer-example5,CNN-Transformer-example6}, use both convolutional and attentional mechanisms to form modules that are iteratively used in the network. No matter which integration method is used, the aim is to allow the model to learn both global and local information\cite{CNN-Transformersurvey}. Previous work in the transparent object depth completion task has demonstrated the importance of the low-level features of the original depth map. This inspired us to design parallel CNN-Transformer structures, which efficiently extract features from the original depth map using separate CNN branches and fuse them with Transformer features, enabling our network to learn both global and local information. 
\section{METHOD}

In this section, we introduce the CNN-Transformer parallel network (TDCNet) for transparent object depth complementation. As shown in Fig. \ref{figure2}, TDCNet consists of two parts, the encoder and the decoder, where the encoder consists of two parallel branches containing the CNN and the Transformer and a fusion structure. The fusion structure collects and fuses features from the dual branches on multiple scales. The MFFM module performs the multi-scale fusion, utilizing spatial attention and channel attention to fully fuse features from different layers. Our decoder consists entirely of convolutional layers and an upsampling module. In addition, we design a new training strategy for model training. Using a damaged depth image of a transparent object region and its complete RGB image, our network generates a comprehensive depth map. Next, we will introduce our parallel structure, fusion method, and loss function in detail, and our training strategy will be introduced with the loss function.

\subsection{CNN-Transformer Parallel Dual-Branch} 
We design a CNN-Transformer parallel backbone for the two branches. Specifically, one branch uses a ResNet-based\cite{resnet} backbone to extract features from the depth map, and the other branch uses a Swin-Transformer\cite{swin} backbone to extract features from the RBG-D features. The advantages of this parallel structure and input setup are that features from the depth map can be extracted by utilizing ResNet's localized feature extraction capability and that the included residual structure ensures the transfer of low-level features to the deeper layers of the network, in addition to the fact that because of the differences in the modeling approaches of CNN and Transformer, the hybrid CNN-Transformer structure has a richer representation of the features compared to the structure composed of a single backbone. 

The CNN branch of TDCNet uses a ResNet18 design with four consecutive ResNet blocks. The input to this branch is a depth map, and the output after the first layer is a feature map with the length and width halved and the number of channels set artificially to C. We set the output feature maps of the remaining layers to half the length and width of the previous layer's output and double the number of channels. 


We designed our Transformer branch to align with the CNN branch, containing four Swin Transformer blocks. The size and number of channels of the feature map output from each layer are aligned with the corresponding layer of the CNN branch. 
   \begin{figure}[tp]
   \vspace*{0.25cm}
      \setlength{\fboxsep}{0pt}%
      \setlength{\fboxrule}{0pt}%
      
      \centering
      \framebox{\includegraphics[width=0.8\linewidth]{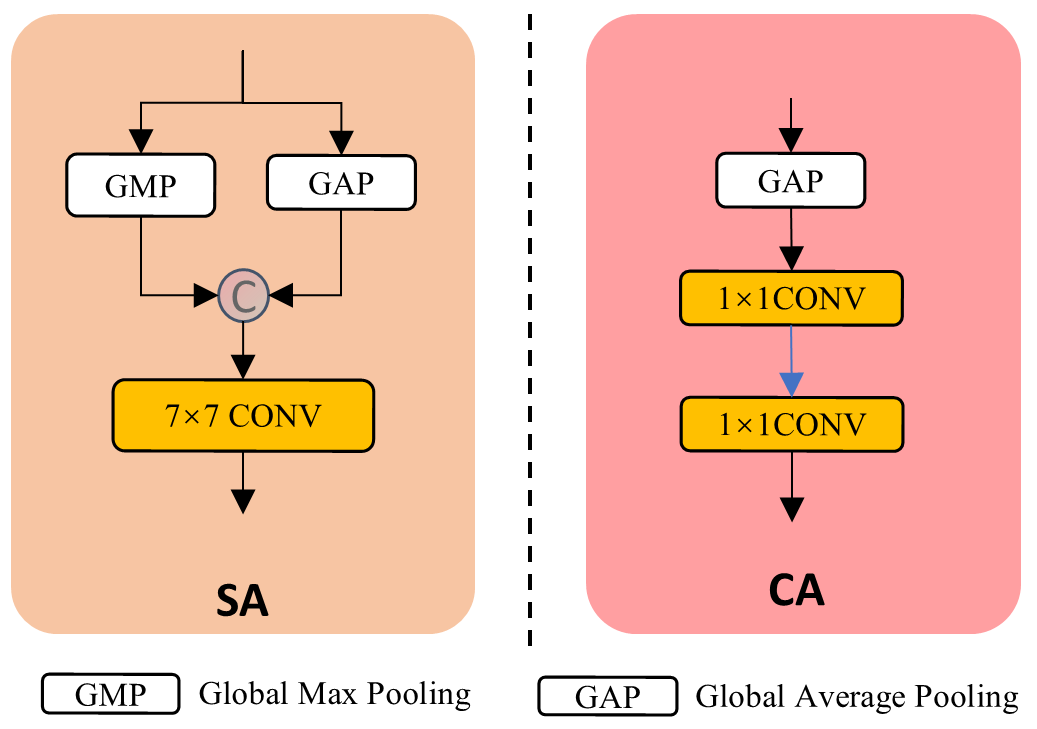}}

      \caption{The structure of SA (spatial attention) and CA (channel attention), where SA and CA compute weight matrices instead of feature maps. The blue arrow on the right represents Relu mapping.}
      \vspace{-0.1cm}
      \label{SA-CA}
   \end{figure}
   
\subsection{Feature fusion}
\textbf{The structure of feature fusion:} As shown in Fig. \ref{network}, our feature fusion structure consists of dual-branch fusion and multi-scale fusion. When reconstructing the depth map in regions where the original depth is trustworthy, reconstruction using features from the CNN branch is easy because the input of the CNN branch is the original depth map. However, regions where the raw depth is not credible rely on high-level features to reconstruct the depth, which may come from the Transformer branch or from the deep output of the CNN branch. To ensure the accurate depth prediction in both depth-credible and non-credible regions, we sum the features from the two branches element by element in order to fuse the features.

After the fusion of multiple layer double branches, we then use the MFFM to do further multi-scale fusion of the fused features. We can describe the overall structure of the feature fusion process as follows:
\setlength\belowdisplayskip{0.2cm}
\begin{equation}
F_{add}(i)=F_{cnn}(i)+F_{trans}\left(i\right)\quad i=1,2,3,4
\end{equation}
\begin{equation}		
		\resizebox{0.9\hsize}{!}{$
		F_{fuse}(i)=\begin{cases}F_{add}(i)&i=1\\ & \\ MFFM(F_{add}(i),F_{fuse}(i-1))&i=2,3,4\end{cases}$
}
\end{equation}

Here $F_{add}(i)$ stands for the dual-branch fusion features of layer $i$, $F_{fuse}(i)$ stands for the multi-scale fusion features of the output of layer $i$, and $MFFM(\text{·})$ stands for our MFFM module.

\textbf{Muti-scale Fusion Module:} We note that fusing features from different layers of the encoder is an effective technique in depth completion tasks\cite{tode,swindrt,fdct}, after passing through many middle layers, low-level features such as edges and contours gradually lose their impact. However, if we enable the fusion of features from the outputs of multiple layers, we can utilize these low-level features at deeper layers. \par

As shown in Fig. \ref{network}, we design a multi-scale feature fusion module MFFM based on channel and space attention. The structure of spatial attention and channel attention is shown in Fig. \ref{SA-CA}. The module adaptively computes weights from features in neighboring layers via channel attention and spatial attention, then uses this weight to fuse features at different scales. Our module first performs adaptive pooling and channel replication on the features of the previous layer, so as to adjust them to the same number of channels and size as the output features of the next layer. Subsequently, the two features are summed and the weights are computed by channel attention and spatial attention. The computation process of our channel attention and spatial attention can be represented as:
\setlength\abovedisplayskip{0.2cm}
\setlength\belowdisplayskip{0.2cm}
\begin{gather}
    X=F_{High}+CR(AAP(F_{Low}))\\
    SA(X)=C_{7\times7}([X_{GMP},X_{GAP}])\\
    CA(X)=C_{1\times1}(max(0,C_{1\times1}(X_{GAP})))
\end{gather}

Where $F_{High}$ and $F_{Low}$ represent features from the shallow level and features from the deep level. $AAP$ stands for adaptive pooling and $CR$ stands for channel replication operation. The length and width of the shallow feature $F_{High}$ are halved in the $AAP$ operation, and then all the channels are copied twice. $X_{GMP}$ and $X_{GAP}$ represent the result of global maximum pooling and global average pooling for $X$, $max(0,\text{·})$ is equivalent to the Relu function.

We sum the two attention weights, connect them to the input features, and shuffle the channels into alternating forms to fully mix the two attentional weights. The two weights are subjected to group convolution and sigmoid function to generate the final weight. This weight is used for weighted summation with features of two scales and skip connections set to mitigate the gradient vanishing problem. Finally, the fused features are projected by a 1 × 1 convolution layer to obtain the final features. This part of the computational process can be represented as:
\setlength\abovedisplayskip{0.2cm}
\begin{gather}
W=\sigma(GConv(CS([SA(X)+CA(X),X])))\\
\resizebox{0.9\hsize}{!}{$
F_{out}=C_{1\times1}(W\times F_{High})+C_{1\times1}((1-W)\times CR(AAP(F_{Low})))$}
\end{gather}

Where $\sigma(\cdot)$ stands for the sigma function, $GConv$ refers to grouped convolution, which contains convolution kernels of size $7\times7$. $CS(\cdot)$ stands for the channel shuffle operation.
\subsection{Loss function}
In addition to the depth loss, some previous work used the smoothing loss as an aid to training based on a priori knowledge of the shape of transparent objects. The formulae for the squared depth loss and smoothed loss are as follows:
\begin{gather}
L_d=||D-\widehat{D}||^2\\
L_{smooth}=1-\cos\langle\hat{V}\times V\rangle 
\end{gather}
Where $D$ and $\widehat{D}$ represent the predicted depth and ground truth, respectively, and $V$ and $\hat{V}$ represent the surface normal vectors calculated from the predicted depth and ground truth.Previous work has often used fixed weighting to combine the two loss functions described above. A common form is shown below:
\setlength\abovedisplayskip{0.2cm}
\begin{gather}
L=L_d+\alpha L_{smooth}
\end{gather}

\begin{table}[bp]
\vspace{-15pt}
\caption{COMPARISON RESULTS ON THE TRANSCG DATASET}
    \resizebox{\linewidth}{!}{
        \begin{tabular}{l|c|c|c|c|c|c}
            \hline
            Methods          & RMSE↓          & REL↓           & MAE↓          & $\sigma_{1.05}\uparrow$ & $\sigma_{1.10}\uparrow$ & $\sigma_{1.25}\uparrow$ \\ 
            \hline
            CG\cite{cg}            & 0.054          & 0.083          & 0.037          & 50.48          & 68.68          & 95.28           \\
            DFNet\cite{transcg}        & 0.018          & 0.027          & 0.012          & 83.76          & 95.67          & 99.71           \\
            LIDF-Refine\cite{implicit} & 0.019          & 0.034          & 0.015          & 78.22          & 94.26          & 99.80           \\
            TCRNet\cite{tcrnet}       & 0.017          & 0.020          & 0.010          & 88.96          & 96.94          & \textbf{99.87}  \\
            TranspareNet\cite{joint} & 0.026          & 0.023          & 0.013          & 88.45          & 96.25          & 99.42           \\
            FDCT\cite{fdct}         & 0.015          & 0.022          & 0.010          & 88.18          & 97.15          & 99.81           \\
            TODE-Trans\cite{tode}   & 0.013          & 0.019          & 0.008          & 90.43          & 97.39          & 99.81           \\
            DualTransNet\cite{segment} & \textbf{0.012} & 0.018          & \textbf{0.008} & \textbf{92.37} & \textbf{97.98} & 99.81           \\
            TDCNet (ours)   & \textbf{0.012} & \textbf{0.017} & \textbf{0.008} & 92.25          & 97.86          & 99.84  \\
            \hline
        \end{tabular}
    }
    
    \label{transcg_comparison}
\end{table}
The $\alpha$ in the above formula is generally set to a fixed value. This fixed weight setting is obviously inflexible, and it is very difficult to select appropriate weights. We believe that the effect of smoothing loss on the depth recovery task changes during the training process. In the early stages of training, the model is able to learn beneficial information from the smoothing loss because the smoothness contains information about the local distribution of depth. And as the value of the smoothing loss approaches convergence, this effect will gradually disappear. And due to the gradient competition between the smoothing loss and the depth loss during the optimization process (which is almost unavoidable), the effect of the smooth loss will even turn negative.

We are inspired by \cite{MTLDeacy,MTLRate}: using both primary and auxiliary losses in the early stages of model training, while focusing on the primary loss in the later stages. In our task, the primary and auxiliary losses correspond to depth loss and smoothing loss, 
respectively. We use the ratio of the first two epochs of loss values to measure the stage of training the task is in. Our final loss function is:
\setlength\belowdisplayskip{0.2cm}
\begin{gather}
R_i(t)=L_i(t-2)/L_i(t-1)\\
\beta=\begin{cases}\alpha& others\\  & \\ 0.1\cdot\alpha & |R_{smooth}(t)|<1.05\end{cases} \label{lossfumla}\\ 
L_{final}=L_d+\beta L_{smooth}
\end{gather}

Where $L_i(t)$ represents the training loss value of the task $i$($smooth$ or $d$) at epoch $t$. 
We use a comparison of $R_i(t)$ and a threshold $1.05$ to measure whether the loss value corresponding to the task $i$ converges. 
The initial value of $\alpha$ is used as a weight when the auxiliary smooth task has not converged, 
and the weight is reduced to one-tenth of the original value when the smooth task is close to convergence.

\section{EXPERIMENT}
\begin{table}[!b]
\vspace*{0.25cm}
\vspace{-15pt}
\caption{COMPARISON WITH STATE-OF-THE-ART METHODS ON CLEARGRASP+OOD DATASETS.}
    \resizebox{\linewidth}{!}{
        \begin{tabular}{l|llllll} 
\hline
Methods              & RMSE↓                 & REL↓           & MAE↓           &$\sigma_{1.05}\uparrow$ & $\sigma_{1.10}\uparrow$ & $\sigma_{1.25}\uparrow$ \\ 
\hline

 \multicolumn{7}{c}{ClearGrasp Real-known}\\ 
\hline
RBG-D-FCN\cite{implicit}        & 0.054                 & 0.087          & 0.048          & 36.32          & 67.11          & 96.26           \\
NLSPN\cite{NLSPN}            & 0.056                 & 0.086          & 0.048          & 40.6           & 67.68          & 96.25           \\
CG\cite{cg}                & 0.032                 & 0.042          & 0.024          & 74.63          & 90.69          & 98.33           \\
LIDF-Refine\cite{implicit}      & 0.028                 & 0.033          & 0.02           & 82.37          & 92.98          & 98.63           \\
TCRNet\cite{tcrnet}           & 0.023                 & 0.027          & 0.016          & \textbf{87.49} & 95.37          & 99.16           \\
TODE-Trans\cite{tode}       & \textbf{0.021}        & \textbf{0.026} & \textbf{0.015} & 86.75          & \textbf{96.59} & 99.25           \\
TDCNet                 & 0.023                 & 0.028          & 0.016          & 86.6           & 95.21          & \textbf{99.8}   \\ 
\hline
 \multicolumn{7}{c}{ClearGrasp Real-novel}\\ 
\hline
RBG-D-FCN\cite{implicit}        & 0.042                 & 0.07           & 0.037          & 42.45          & 75.68          & 99.02           \\
NLSPN\cite{NLSPN}             & 0.036                 & 0.059          & 0.03           & 51.97          & 84.82          & 99.52           \\
CG\cite{cg}               & 0.027                 & 0.039          & 0.022          & 79.5           & 93             & 99.28           \\
LIDF-Refine\cite{implicit}      & 0.025                 & 0.036          & 0.02           & 76.21          & 94.01          & 99.35           \\
TCRNet\cite{tcrnet}          & 0.014                 & 0.025          & 0.012          & 81.93          & 95.12          & 99.67           \\
TODE-Trans\cite{tode}        & 0.012                 & 0.016          & \textbf{0.008} & 95.74          & 99.08          & 99.96           \\
TDCNet                 & \textbf{0.011}        & \textbf{0.015} & \textbf{0.008} & \textbf{96.01} & \textbf{99.32} & \textbf{99.97}  \\ 
\hline
 \multicolumn{7}{c}{ClearGrasp Sys-novel}\\ 
\hline
RBG-D-FCN\cite{implicit}        & 0.033                 & 0.058          & 0.028          & 52.4           & 85.64          & 98.94           \\
NLSPN\cite{NLSPN}            & 0.029                 & 0.049          & 0.024          & 64.83          & 88.2           & 98.57           \\
CG\cite{cg}               & 0.037                 & 0.062          & 0.032          & 50.27          & 84             & 98.39           \\
LIDF-Refine\cite{implicit}     & 0.028                 & 0.048          & 0.023          & 68.62          & 89.1           & 99.2            \\
TCRNet\cite{tcrnet}           & 0.023                 & 0.04           & 0.018          & 71.33          & 90.84          & \textbf{99.49}  \\
TDCNet                 & \textbf{0.021}        & \textbf{0.034} & \textbf{0.017} & \textbf{79.49} & \textbf{93.4}  & 99.1            \\ 
\hline
 \multicolumn{7}{c}{ClearGrasp Sys-known}\\  
\hline
RBG-D-FCN\cite{implicit}       & 0.028                 & 0.039          & 0.021          & 76.53          & 91.82          & 99              \\
NLSPN\cite{NLSPN}           & 0.026                 & 0.041          & 0.021          & 74.89          & 89.95          & 98.59           \\
CG\cite{cg}               & 0.034                 & 0.045          & 0.026          & 73.53          & 92.68          & 98.25           \\
LIDF-Refine\cite{implicit}     & 0.012                 & 0.017          & 0.009          & 94.79          & 98.52          & 99.67           \\
TCRNet\cite{tcrnet}         & 0.01                  & 0.015          & \textbf{0.006} & \textbf{95.83} & 98.74          & 99.75           \\
TDCNet                 & \textbf{0.009}        & \textbf{0.014} & 0.007          & 95.32          & \textbf{98.89} & \textbf{99.77}  \\
\hline
\end{tabular}
    }
    \label{CG_comparison}
\end{table}
\subsection{Datasets and metrics}
\textbf{Datasets:} The datasets we use for training and evaluation include TransCG\cite{transcg}, ClearGrasp\cite{cg}, and Omniverse\cite{odd}. The training samples in the ClearGrasp dataset were generated by the Synthesis AI platform, which consists of nine CAD models using transparent plastic objects from the real world to generate a total of 23524 datasets for training. The test data for ClearGrasp consists of both synthetic and real-world data with transparent object shapes that are both present in the training data and not included in the training data.TransCG uses a robot to collect a new dataset based on a real environment that contains 57,715 RGB-D images. The dataset contains 51 objects that are common in everyday life and may cause inaccuracies in depth images, including reflective objects, transparent objects, translucent objects, and objects with dense holes. The Omniverse (ODD) dataset contains approximately 68,000 samples of training data based on simulated environments and consists of a diverse collection of transparent and opaque objects randomly placed in a cluttered scene in a simulation. The objects are randomly placed in cluttered scenes within the simulation.

\textbf{Metrics:} We choose the root mean square error (RMSE), the absolute relative difference (REL), and the mean absolute error (MAE) as evaluation metrics.

The RMSE is a common metric used to assess the quality of the predicted depth images and is defined as:$\sqrt{\frac{1}{|\widehat{D}|}\sum_{d\in\widehat{D}}\|d-d^*\|^2}$. $d$ and $d^*$ represent the prediction depth and ground truth.

REL is the mean absolute relative error between the predicted depth and the ground truth, defined as:$\frac1{|\widehat{\mathcal{D}}|}\Sigma_{d\in\widehat{\mathcal{D}}}|d-d^*|/d^*$.

MAE is the mean absolute error between the predicted depth and the ground truth: $\frac1{|\widehat{D}|}\sum_{d\in\widehat{D}}|d-d^*|$.

Finally, we calculate the percentage of pixels whose predicted depth satisfies $\max\left(\frac{d_{i}}{d_{i}^{*}},\frac{d_{i}^{*}}{d_{i}}\right)<\delta$, where $delta$ is set by us to 1.05, 1.10, and 1.25.

   \begin{figure}[!t]
   \vspace*{0.25cm}
      \setlength{\fboxsep}{0pt}%
      \setlength{\fboxrule}{0pt}%
      \flushright
      \framebox{\includegraphics[width=0.95\linewidth]{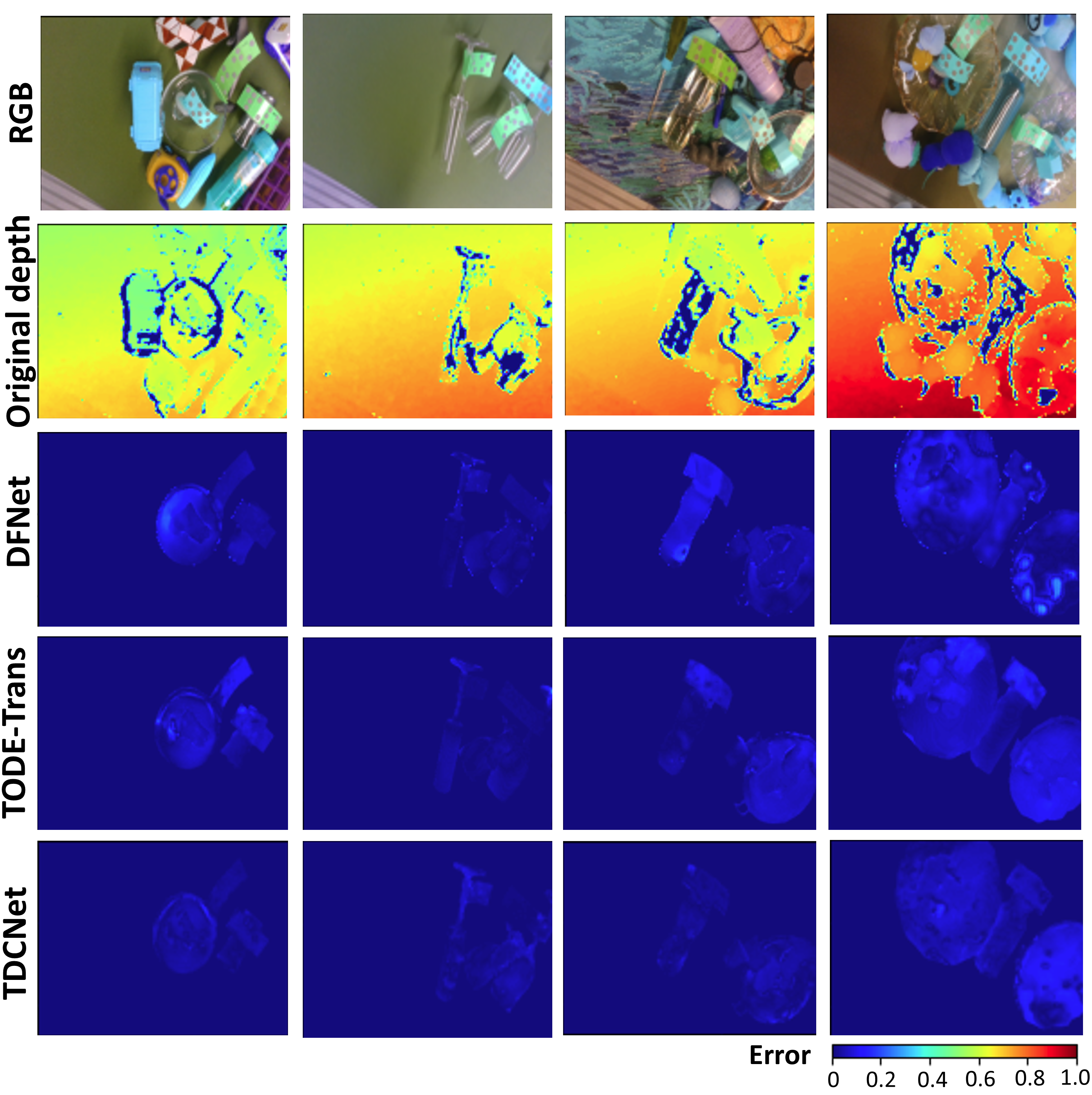}

}    
      \caption{\centering{
     The visualization result on TransCG dataset. Each pixel of the error map is calculated by the following relative error: $|d-d^* |/d^*$.}
      }
      \label{TransCG}
      \vspace{-15pt}
   \end{figure}

   \begin{figure}[htbp]
   \vspace*{0.25cm}
      \setlength{\fboxsep}{0pt}%
      \setlength{\fboxrule}{0pt}%
      \flushright
      \framebox{\includegraphics[width=0.95\linewidth]{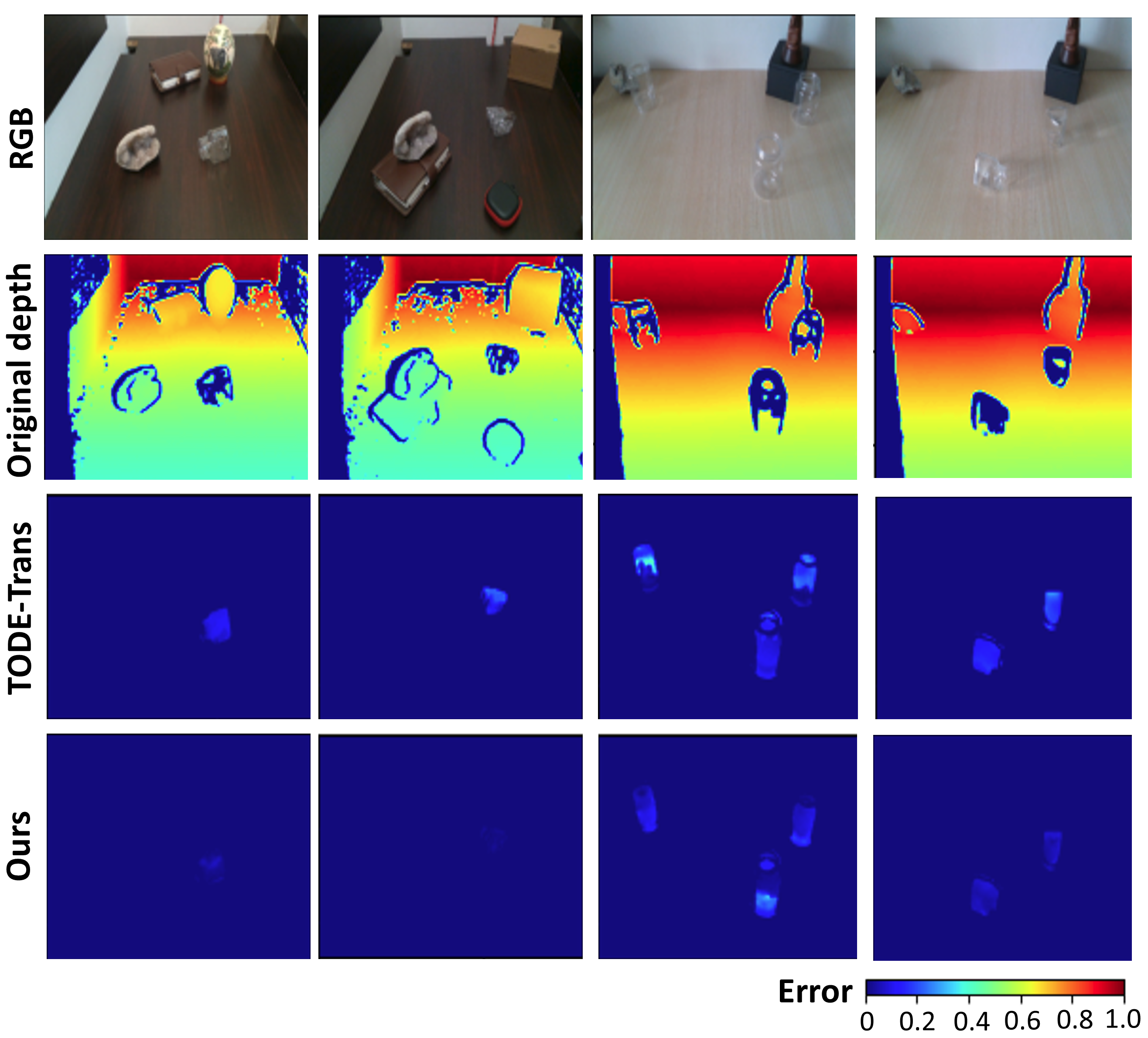}

}
      \caption{
     The visualization result on ClearGrasp dataset. Each pixel of the error map is calculated by the following relative error: $|d-d^* |/d^*$.
      }
      \label{CG}
      \vspace{-10pt}
   \end{figure}
\subsection{Implementation Details}

All our experiments were performed on Intel I9-14900k CPU and Nvidia RTX 4090 GPU. During training we set the batch size to $16$ and the input image size to $320\times240$. The value of $\alpha$ in loss function Eq. \ref{lossfumla} is set to $0.1$. We trained using the AdamW optimizer with an initial learning rate of $1e-3$ and decayed the learning rate to one-tenth of the learning rate every $15$ rounds for a total of $40$ rounds of training. We used a variety of data enhancement methods prior to training, including random flipping, rotation, and random noise. The number of channels C in Fig. \ref{figure2} was set to $24$.

\begin{table}
\centering
\caption{Generalization Test}
\resizebox{\linewidth}{!}{
\begin{tblr}{
  cell{1}{1} = {r=2}{},
  cell{1}{2} = {c=6}{},
  vline{2} = {1,3-9}{},
  hline{1,3,10} = {-}{},
  hline{2} = {2-7}{},
}
Methods            & Train ClearGrasp$+$OOD Val TransCG     ~ &                &                &               &                &                \\
                   & RMSE$\downarrow $                                 & REL$\downarrow $          & MAE$\downarrow$           & $\sigma1.05$$\uparrow$        & $\sigma1.10$$\uparrow$          & $\sigma1.25$$\uparrow$          \\
CG\cite{cg}              & $0.061$                                  & $0.108$          & $0.049$          & $33.59$         & $54.73$          & $92.48$          \\
LIDF$-$Refine\cite{implicit}    & $0.146$                                  & $0.262 $         & $0.115$          & $13.7$          & $26.39 $         & $57.95$          \\
DFNet\cite{transcg}          & $0.048$                                  & $0.088$          & $0.039 $         & $38.65$         & $64.42 $         & $95.28$          \\
TransparentNet\cite{joint} & $0.071$                                  & $0.06$           & $0.036$          & $62.99 $        & $82.92$          & $95.93$          \\
TODE$-$Trans\cite{tode}     & $0.034 $                                 & $0.057 $         & $0.026 $         & $64.1$          & $78.86$          & $98.8 $          \\
TCRNet\cite{tcrnet}         & $0.033$                                  & $0.055 $         & $0.023$          &$ \textbf{64.7}$ & $81.75$          & $99.32$  \\
TDCNet               & $\textbf{0.027}$                         & $\textbf{0.048}$ & $\textbf{0.021}$ & $61.73 $        & $\textbf{88.33}$ & $\textbf{99.48} $
\end{tblr}
}
\label{generalization}
\vspace{-15pt}
\end{table}

\subsection{Depth Completion Experiments}
We compare with current state-of-the-art methods on the TransCG dataset and the ClearGrasp dataset. Following previous works\cite{fdct,tode,tcrnet,tim}, we train the ClearGrasp dataset by adding the OOD dataset along with it, and then test network on the ClearGrasp test dataset. The quantitative comparison results are shown in Table \ref{transcg_comparison} and Table \ref{CG_comparison}. 
Our model achieves state-of-the-art performance on both datasets and outperforms previous single-branch networks. Our TDCNet requires only RGB and depth image inputs for training, but we outperform the previous DualTransNet\cite{segment} and FDCT\cite{fdct} in terms of key performance metrics, which require additional inputs or network for training.

In Fig. \ref{TransCG} and Fig. \ref{CG}, we show qualitative comparisons between our model and other models on the TransCG and ClearGrasp datasets. The error map shown in Fig. \ref{TransCG} and Fig. \ref{CG} is only the part of transparent object. It is clear that our dual-branch model is able to obtain more accurate predictions in regions where there are plausible depths or edges in the original depth map.

Finally, to exemplify the generalization ability of TDCNet, we trained it with ClearGrasp and OOD datasets and tested it on TransCG. The results of our generalization experiments are shown in Table \ref{generalization}. TDCNet exhibits exceptional generalization capability owing to its utilization of the input depth map.

\subsection{Ablation study}
\textbf{1) CNN-Transformer parallel dual-branch structure:} We show the effectiveness of our CNN-Transformer parallel dual-branch structure in two ways. Firstly, we replace the backbone of the two branches to demonstrate the impact of the CNN-Transformer parallelised backbone on our network. Table \ref{backbone} is the outcome of our experiment on the ClearGrasp dataset. Table \ref{backbone} verifies the advantages of our CNN-Transformer parallel backbone over other single backbones, where the third model is equivalent to exchanging the inputs of two branches with each other. Next, we experiment with various inputs during the dual-branch process to demonstrate the impact of our input selections. As shown in Table \ref{input}, we also conduct experiments on the ClearGrasp dataset to evaluate the impact of our input selections. And Table \ref{input} verifies that CNN's feature extraction model is more suitable for extracting features from depth maps, while Transformer is more suitable for RBG-D images.

\begin{table}
\vspace*{0.25cm}
\centering
\caption{Ablation experiments with dual-branch backbone selection}
\resizebox{\linewidth}{!}{
\begin{tblr}{
  cell{1}{1} = {c=8}{},
  cell{2}{1} = {r=2}{},
  cell{2}{2} = {c=2}{},
  cell{2}{4} = {c=2}{},
  cell{2}{6} = {r=2}{},
  cell{2}{7} = {r=2}{},
  cell{2}{8} = {r=2}{},
  vline{2} = {2,4-7}{},
  hline{1-2,4,8} = {-}{},
  hline{3} = {2-5}{},
}
train ClearGrasp$+$ODD val ClearGrasp Real$-$novel &           &            &           &            &                &                &                \\
Models                                         & backboneA &            & backboneB &            & RMSE↓          &REL↓           & MAE↓           \\
                                               & CNN       & Swin$-$Trans & CNN       & Swin$-$Trans &                &                &                \\
1                                              & $\checkmark$        &            & $\checkmark $        &            & 0.017          & 0.03           & 0.018          \\
2                                              &           & $\checkmark$          &           & $\checkmark$          & 0.012          & 0.016          & 0.009          \\
3                                              &           & $\checkmark $         & $\checkmark$         &            & 0.013          & 0.018          & 0.01           \\
4(ours)                                        & $\checkmark$         &            &           & $\checkmark$          & \textbf{0.011} & \textbf{0.015} & \textbf{0.008} 
\end{tblr}
}
\label{backbone}
\end{table}

\begin{table}[!t]
\vspace{-10pt}
\centering
\caption{Ablation experiments with dual-branch input selection}
\resizebox{\linewidth}{!}{
\begin{tblr}{
  cell{1}{1} = {c=6}{},
  vline{2} = {2-6}{},
  hline{1-3,7} = {-}{},
}
train ClearGrasp+ODD val ClearGrasp Real-novel &                    &                    &                                            &                                           &                                           \\
Models                                         & input of backboneA & input of backboneB & RMSE↓ & REL↓ & MAE↓\\
1                                              & RGB-D              & RGB-D              & 0.012                                      & 0.017                                     & 0.009                                     \\
2                                              & Depth              & RGB                & 0.021                                      & 0.033                                     & 0.017                                     \\
3                                              & RGB-D              & RGB                & 0.013                                      & 0.018                                     & 0.01                                      \\
4(ours)                                        & Depth              & RGB-D              & \textbf{0.011}                             & \textbf{0.015}                            & \textbf{0.008}                            
\end{tblr}
}
\label{input}
\end{table}

\textbf{2) Fusion Module:} We conduct ablation experiments on the fusion module MFFM that TDCNet utilizes. We make comparisons between TDCNet that removes multi-scale fusion and TDCNet that uses FFM from \cite{tode}. FFM accomplishes fusion by directly multiplying feature maps from the previous level with those from the next level using adjusted sizes and channel counts as weights. In contrast, our MFFM module calculates channel and spatial attention as weights based on the input feature. Table \ref{CSAF} demonstrates the effectiveness of our multiscale fusion and MFFM modules.

\begin{table}[!t]
\vspace{-0.25cm}
\centering
\caption{Ablation experiments with the MFFM module on the TransCG dataset}
\resizebox{\linewidth}{!}{
\begin{tblr}{
  vline{2} = {-}{},
  hline{1-2,5} = {-}{},
}
Methods     & RMSE↓          & REL↓           & MAE↓           & $\sigma1.05\uparrow $        & $\sigma1.10\uparrow $        & $\sigma1.25\uparrow$         \\
ours         & 0.014          & 0.02           & 0.009          & 89.73          & 97.1           & 99.78          \\
ours+FFM\cite{tode} & 0.014          & 0.019          & \textbf{0.008} & 91.03          & 97.24          & 99.74          \\
ours+MFFM    & \textbf{0.013} & \textbf{0.018} & \textbf{0.008} & \textbf{91.54} & \textbf{97.88} & \textbf{99.81} 
\end{tblr}
\label{CSAF}
}
\vspace{-15pt}
\end{table}

\textbf{3) Loss function:} We perform ablation experiments of our training strategy on the TransCG dataset. The experimental results presented in Table \ref{loss} demonstrate that our weight decay strategy is more effective compared to fixed weights.

\begin{table}
\vspace*{0.25cm}
\centering
\caption{Ablation experiments with the loss function on the TransCG dataset}
\resizebox{\linewidth}{!}{
\begin{tblr}{
  vline{2} = {-}{},
  hline{1-2,6} = {-}{},
}
Methods        & RMSE↓          & REL↓           & MAE↓           & $\sigma1.05\uparrow$         & $\sigma1.10\uparrow$         & $\sigma1.25\uparrow$         \\
$L_d$ & 0.013          & 0.019          & 0.009          & 91.23          & 97.73          & 99.77          \\
$L_d+0.1L_{smooth}$    & 0.013          & 0.018          & \textbf{0.008} & 91.24          & 97.78          & 99.81          \\
$L_d+0.01L_{smooth}$   & 0.013\textbf{} & 0.019\textbf{} & 0.009\textbf{} & 90.39\textbf{} & 97.67\textbf{} & \textbf{99.86} \\
ours & \textbf{0.012} & \textbf{0.017} & \textbf{0.008} & \textbf{92.25} & \textbf{97.86} & 99.84\textbf{} 
\end{tblr}
}
\label{loss}
\vspace{-10pt}
\end{table}

\subsection{Analysis}
Our network uses a parallel dual-branch CNN-Transformer structure to extract the RBG-D and the only depth features separately, thus alleviating the lack of utilization of the original depth map by previous methods. As shown in Table \ref{transcg_comparison}, our performance on the TransCG dataset outperforms almost all previous all single-branch models. It even outperforms segment containing subnetworks in major metrics. We also demonstrate state-of-the-art performance on the ClearGrasp dataset.
 To further demonstrate the effectiveness of our dual-branch structure, we designed ablation experiments for the backbone and inputs of the dual-branch. The results in Table \ref{input} demonstrate that our dual-branch input setup is superior to the other setups. Meanwhile, Table \ref{backbone} reflects the complementary advantages of our CNN-Transformer structure.

As shown in Table \ref{generalization}, our model performs well on the generalization experiments. We attribute this to the fact that TDCNet uses a ResNet-based backbone and a fusion structure that contains a large number of residual connections, thus greatly increasing our generalization ability.

We use the MFFM module to fuse multi-scale features. Compared to the way of using shallow feature maps as weights in \cite{tode}, calculating weights of MFFM in both the spatial and channel dimensions of the feature maps is more flexible and efficient, as verified by the results of the ablation experiments in Table \ref{CSAF}.

We argue that the positive impact of the smoothing loss during network training is not permanent. So our training strategy is to suppress the weights of the smoothing loss as it approaches convergence. The experiments in Table \ref{loss} demonstrate that our strategy outperforms the strategy using fixed weights.

\section{Conclusions}
In this paper, we propose a dual-branch network for transparent object depth completion. Our network adopts an encoder-decoder design with a CNN-Transformer parallel backbone. The dual-branching extracts features from the original depth map and RBG-D image, respectively, and thus achieves accurate depth prediction in both heavily depth-impaired and less depth-impaired regions. Experimental results show that due to our dual-branch design, the model is able to accurately complement the depth-missing regions while maintaining plausible edges in the original depth map. This capability is beneficial for downstream applications such as robot grasping.



\bibliography{reference}

\end{document}